\documentclass[letterpaper]{article} 
\usepackage[draft]{aaai25}  
\usepackage{times}  
\usepackage{helvet}  
\usepackage{courier}  
\usepackage[hyphens]{url}  
\usepackage{graphicx} 
\usepackage{amsmath}
\usepackage{amsthm}
\usepackage{amsfonts}
\usepackage{multirow}
\usepackage{booktabs}
\usepackage{graphicx} 
\usepackage{xcolor}   

\usepackage{array}

\newtheorem{assumption}{Assumption}
\newtheorem{remark}{Remark}
\newtheorem{theorem}{Theorem}
\newtheorem{example}{Example}
\usepackage{natbib}  
\usepackage{caption} 
\usepackage{algorithm}
\usepackage{algorithmic}

\usepackage{newfloat}
\usepackage{listings}
\DeclareCaptionStyle{ruled}{labelfont=normalfont,labelsep=colon,strut=off} 
\floatstyle{ruled}
\newfloat{listing}{tb}{lst}{}
\floatname{listing}{Listing}
\title{Domain Adaptive Unfolded Graph Neural Networks}
\author{
    Zepeng Zhang,
    Olga Fink
}
\affiliations{
    Intelligent Maintenance and Operations Systems (IMOS) Lab

    \'{E}cole Polytechnique F\'ed\'erale de Lausanne (EPFL), Lausanne, Switzerland\\
    zepeng.zhang@epfl.ch, olga.fink@epfl.ch
}

\usepackage{bibentry}

\begin{document}

\maketitle

\begin{abstract}
Over the last decade, graph neural networks (GNNs) have made significant progress in numerous graph machine learning tasks.
In real-world applications, where domain shifts occur and labels are often unavailable for a new target domain, graph domain adaptation (GDA) approaches have been proposed to facilitate knowledge transfer from the source domain to the target domain.
Previous efforts in tackling distribution shifts across domains have mainly focused on aligning the node embedding distributions generated by the GNNs in the source and target domains.
However, as the core part of GDA approaches, the impact of the underlying GNN architecture has received limited attention.
In this work, we explore this orthogonal direction, i.e., how to facilitate GDA with architectural enhancement.
In particular, we consider a class of GNNs that are designed explicitly based on optimization problems, namely unfolded GNNs (UGNNs), whose training process can be represented as bi-level optimization.
Empirical and theoretical analyses demonstrate that when transferring from the source domain to the target domain, the lower-level objective value generated by the UGNNs significantly increases, resulting in an increase in the upper-level objective as well.
Motivated by this observation, we propose a simple yet effective strategy called cascaded propagation (CP), which is guaranteed to decrease the lower-level objective value.
The CP strategy is widely applicable to general UGNNs, and we evaluate its efficacy with three representative UGNN architectures.
Extensive experiments on five real-world datasets demonstrate that the UGNNs integrated with CP outperform state-of-the-art GDA baselines.
\end{abstract}

\section{Introduction}
Over the past decade, graph neural networks (GNNs) have become increasingly prevalent for processing graph-structured data, achieving significant advancements across a wide range of applications, including social networks, engineering applications, molecular chemistry,  recommender systems, and more \cite{wu2020comprehensive, zhou2020graph, GNNBook2022}.
Despite their achievements, the majority of GNNs rely on supervised training, which requires a substantial amount of labeled data.
This reliance limits their applicability in real-world applications where domain shifts occur and labels for a new target domain are often unavailable \cite{liu2023structural}.
To mitigate such issues, a variety of graph domain adaptation (GDA) approaches have been developed \cite{shi2024graph, cai2024graph}.
GDA aims to transfer knowledge from a labeled source domain to an unlabeled target domain, which often experiences distribution shifts.
The main challenge of GDA is to address various domain shifts in node features, structural patterns, and task labels, which typically hinder knowledge transfer.

Inspired by the success of domain adaptation (DA) in sequence and image data \cite{ganin2015unsupervised,wang2018deep}, some early attempts in GDA have employed  DA techniques developed for computer vision tasks to learn domain-invariant and label-discriminative embeddings \cite{zhang2021adversarial, shen2021network,dai2022graph}.
However, these approaches have overlooked the unique characteristics of graph-structured data.
Subsequently, researchers have begun to examine the distribution shifts that are particularly prevalent in graph-structured data.
For example, \citet{guo2023learning} investigate the influence of node degree distribution shifts, \citet{shi2023improving} study the shifts in hierarchical graph structure, \citet{wu2023non} analyze the graph subtree discrepancies, and \citet{liu2023structural} explore the effect of conditional structure shifts. 

While these techniques have significantly advanced the field of  GDA, they primarily focus on aligning the distributions of node embeddings between the source and target domains to mitigate the impact of distribution shifts \cite{liu2023structural,zhu2023explaining}.
In computer vision, recent work has begun to examine the effect of the underlying architecture, which has been shown to play a critical role in DA \cite{li2023sparse}.
However, the effect of the underlying architecture has been largely overlooked in the field of GDA.
Some recent works have started to investigate how different components in message-passing GNNs affect domain generalization (DG) \cite{guo2024investigating,liu2024rethinking}.
Their results show that both the decoupled structure and the attention mechanism contribute positively to DG, while the transformation layers impair the generalization capability.
Based on these findings, \citet{liu2024rethinking} propose a GNN model with a decoupled structure, while \citet{guo2024investigating} further adopts the attention mechanism in the model.
However, in contrast to the majority of GDA methods that can be applied to various GNN architectures, \citet{guo2024investigating,liu2024rethinking} only propose two particular GNNs, indicating limitations in 
extending their approaches to other GNNs and addressing issues with distinctive properties like heterophily and heterogeneity.
This limited applicability naturally leads to an important question: \textit{Can we facilitate GDA across a wide spectrum of GNNs through architectural enhancement?}
Since there is currently no framework encompassing all GNNs, we focus on a subset of GNNs known for their relatively high transparency, namely unfolded GNNs (UGNNs) \cite{yang2021graph,ma2021unified,zhu2021interpreting}.
It is important to highlight that our focus on UGNNs encompasses a broad spectrum of models, including several widely adopted GNN architectures such as the approximate personalized propagation of neural predictions (APPNP) model \cite{gasteiger2019predict}, the Generalized PageRank GNN (GPRGNN) model \cite{chien2021adaptive}, and the elastic GNN (ElasticGNN) model \cite{liu2021elastic}.
Unlike conventional GNNs, UGNNs are rigorously derived from iterative optimization algorithms for solving optimization problems.
Thus, the training of a UGNN can be represented as solving a bi-level optimization problem \cite{zheng2024bloomgml}, where the optimal solution of a lower-level problem serves as input features for an upper-level loss minimization problem.
Benefiting from the transparent nature of UGNNs, we can integrate desired inductive biases into the model by modifying the lower-level problem associated with UGNNs.
We can also better understand the behavior of UGNNs by analyzing their corresponding bi-level optimization problems.

In this study, we first perform empirical investigations to understand how UGNNs perform when transferring from the source to the target domain.
We begin our analyses with a simple UGNN model: APPNP \cite{gasteiger2019predict}, whose lower-level objective corresponds to graph signal denoising (GSD) \cite{ma2021unified,zhu2021interpreting}.
We empirically evaluate how the lower-level objective for APPNP changes when transferring from the source domain to the target domain and observe that the GSD objective value in the target domain is significantly larger than in the source domain.
This interesting observation suggests that the increase in the lower-level objective may be related to the increase in the upper-level objective, i.e., the loss function.
To better understand how general UGNNs behave when transferring from the source domain to the target domain, we theoretically analyze how the two objectives in bi-level optimization change for general UGNNs.
Based on our empirical and theoretical analyses, we propose a simple yet effective strategy called \textbf{Cascaded Propagation} (CP) to enhance the DG ability of UGNNs.
Specifically, CP involves reinjecting the output of the lower-level problem back as its input. 
By applying the CP strategy, the lower-level objective is provably decreased, which potentially leads to a reduction in the upper-level loss function as well.
We then showcase how the proposed strategy works with three different UGNN architectures, namely APPNP \cite{gasteiger2019predict}, GPRGNN  \cite{chien2021adaptive}, and ElasticGNN \cite{liu2021elastic}.
It is worth highlighting that since the proposed CP strategy enhances UGNNs at the architectural level, node distribution aligning-based methods can be employed together to further improve the effectiveness in tackling GDA tasks.
Our contributions can be summarized as follows:
\begin{itemize}
    \item We investigate both empirically and theoretically how the values of the upper- and lower-level objectives in bi-level optimization, associated with UGNNs, change when transferring from the source to the target domain.  
     The results indicate that there is a significant increase in lower-level objectives when GDA is performed between the training and testing processes.
    \item Based on our empirical and theoretical analyses, we propose a simple yet effective method called Cascaded Propagation (CP).
    It builds upon theory-guided DG enhancement for UGNNs and is guaranteed to decrease the lower-level objective value.
    \item We integrate three UGNNs with the proposed CP and evaluate their performance on GDA tasks. 
    Experiments on 8 real-world node classification GDA tasks demonstrate the effectiveness of the proposed strategy,  providing substantial and consistent performance improvement across various UGNN architectures and datasets.
\end{itemize}

\section{Preliminaries and Background}
\textbf{Notations.} We use $\mathcal{G}=(\mathcal{V},\mathcal{E})$ to denote an unweighted graph with $\mathcal{V}$ and $\mathcal{E}$ being the node set and the edge set, respectively. 
The graph's adjacency matrix is given by $\mathbf{A}\in\mathbb{R}^{N\times N}$. 
We denote by $\mathbf{1}$ and $\mathbf{I}$ the all-one column vector and the identity matrix, respectively.
Given $\mathbf{D}=\mathrm{Diag}\left(\mathbf{A}\mathbf{1}\right)\in\mathbb{R}^{N\times N}$ as the diagonal degree matrix, the Laplacian matrix is defined as $\mathbf{L}=\mathbf{D}-\mathbf{A}$.
We denote by $\mathbf{A}_{\rm sym}=\mathbf{D}^{-\frac{1}{2}}\mathbf{A}\mathbf{D}^{-\frac{1}{2}}$ the symmetric normalized adjacency matrix. 
Subsequently, the symmetric normalized Laplacian matrix is defined as $\mathbf{L}_{\rm sym}=\mathbf{I}-\mathbf{D}^{-\frac{1}{2}}\mathbf{A}\mathbf{D}^{-\frac{1}{2}}$.
In GNNs, the symmetric normalized adjacency matrix and Laplacian matrix are commonly used. 
Thus, in the following, we will use $\mathbf{A}$ and $\mathbf{L}$ to represent their normalized counterparts for simplicity.
$\hat{\mathbf{X}}\in\mathbb{R}^{N\times M}$ is a node feature matrix or a graph signal matrix, with $M$ being the dimension of the node feature.
$\mathbf{Y}\in\mathbb{R}^{N\times C}$ is a label matrix for node classification tasks with $C$ being the number of classes.

\subsection{Graph Domain Adaptation}
Unsupervised GDA (UGDA) aims to transfer the knowledge learned from a labeled source graph domain to an unlabeled target graph domain with potential domain gaps between the two graph domains \cite{shi2024graph}.
We denote the source graph domain as $\mathcal{G}^{s}=\left\{ \mathcal{V}^{s},\mathcal{E}^{s},\hat{\mathbf{X}}^{s},\mathbf{Y}^{s}\right\} $ and the target graph domain as $\mathcal{G}^{t}=\left\{ \mathcal{V}^{t},\mathcal{E}^{t},\hat{\mathbf{X}}^{t},\mathbf{Y}^{t}\right\} $, where $\mathbf{Y}^s$ and $\mathbf{Y}^t$ represent the labels.
Note that in the setting of GDA, $\mathbf{Y}^t$ is unavailable during training.
Typically, the source and target domains share the same task and node feature space. Examples include citation networks from different periods \cite{zhang2021adversarial}, social networks from different platforms \cite{wang2023cross}, and proteins from various species \cite{you2023graph}.
However, distribution shifts may still occur in the generated node embeddings and the topological attributes \cite{guo2023learning,you2023graph,shi2023improving}.
The objective of GDA is to train a model using information from the source graph $\mathcal{G}^s$, along with $\mathcal{V}^{t},\mathcal{E}^{t},\hat{\mathbf{X}}^{t}$, to accurately predict the labels in the target domain $\mathbf{Y}^{t}$.

Most existing UGDA methods focus on minimizing domain discrepancy in the node representation space \cite{shi2024graph}.
For example, \cite{shen2021network} and \cite{wu2023non} train the model to learn domain-invariant representations by minimizing pre-defined domain discrepancy metrics such as maximum mean discrepancy (MMD) \cite{gretton2012kernel} and graph subtree discrepancy \cite{wu2023non}.
Besides explicitly minimizing domain discrepancies, some approaches use adversarial training techniques that employ a domain classifier to predict the domain from which the representation is generated \cite{wu2020unsupervised,shen2020adversarial,dai2022graph}.
However, the influence of the GNN architecture on GDA has been significantly underexplored.
Recently, efforts have been made to investigate the impact of different components in message passing GNNs on DG \cite{guo2024investigating} and DA capabilities \cite{liu2024rethinking}.
However, these studies have primarily focused on developing specific GNN architectures. A broadly applicable enhancement that can be integrated across a wider range of GNNs remains lacking.

\subsection{Unfolded Graph Neural Networks}
In recent years, research has shown that the message passing layers in various GNNs could often be understood as gradient steps for solving specific optimization problems \cite{yang2021graph,ma2021unified,zhu2021interpreting,zhang2022towards}. 
This interpretation applies to several models, such as GCN \cite{kipf2017semi}, APPNP \cite{gasteiger2019predict}, and GPRGNN \cite{chien2021adaptive}, among others.
For example, in the APPNP  model \cite{gasteiger2019predict}, the message passing scheme can be described using the  input $\mathbf{X}$ and adjacency matrix $\mathbf{A}$ as follows: 
\begin{equation}
\begin{aligned}
    &\mathbf{H}^{(0)}=\mathbf{X}=\mathsf{MLP}(\hat{\mathbf{X}}),\ \ \ \ \\ &\mathbf{H}^{(k+1)}=\left(1-\alpha\right)\mathbf{A}\mathbf{H}^{(k)}+\alpha\mathbf{X}, \ \text{for}\ k=0,\ldots,K-1,
\end{aligned}
\label{eq:APPNP}
\end{equation}
where $\mathbf{H}^{(k)}$ represents the learned feature after the $k$-th layer and $\alpha$ is the teleport probability.
From an optimization perspective, the process defined in Eq. \eqref{eq:APPNP} can be seen as executing $K$ steps of gradient descent to solve the following problem with initialization $\mathbf{H}^{(0)}=\mathbf{X}$ and step size $0.5$ \cite{zhu2021interpreting,ma2021unified,zhang2022towards}:
\begin{equation}
\begin{aligned} & \underset{\mathbf{H}\in\mathbb{R}^{N\times M}}{\mathsf{minimize}} &  & \alpha\left\Vert \mathbf{H}-\mathbf{X}\right\Vert _{{\rm F}}^{2}+\left(1-\alpha\right)\mathrm{Tr}\bigl(\mathbf{H}^{\top}\mathbf{L}\mathbf{H}\bigr),\end{aligned}
\label{eq:GSD}
\end{equation}
where $\mathbf{X}$ and $\alpha$ share the same meaning as in Eq. \eqref{eq:APPNP}. 
Problem \eqref{eq:GSD} carries the meaning of GSD, where the first term is a fidelity term representing the distance between the recovered graph signal $\mathbf{H}$ and the noisy graph signal $\mathbf{X}$, and the second term is the Laplacian smoothing term measuring the variation of the recovered graph signal $\mathbf{H}$.

Based on the optimization interpretation of UGNNs, their training process can be further represented as bi-level optimization \cite{zhang2022towards,zheng2024bloomgml}:
\begin{equation}
\begin{aligned} & \underset{\boldsymbol{\Theta}_{\mathsf{pre}},\boldsymbol{\Theta}_{\mathsf{pos}}}{\mathrm{min}} &  & f_{\mathsf{up}}\left(\mathbf{Y},p_{\mathsf{pos}}(\bar{\mathbf{X}},\boldsymbol{\Theta}_{\mathsf{pos}})\right) & \mathsf{(upper)}\\
 & \;\:\mathrm{s.t.} &  & \bar{\mathbf{X}}\in\mathrm{arg}\min_{\mathbf{H}}f_{\mathsf{low}}\left(\mathbf{H},\mathbf{L},p_{\mathsf{pre}}(\hat{\mathbf{X}},\boldsymbol{\Theta}_{\mathsf{pre}})\right), & \mathsf{(lower)}
\end{aligned} 
\label{eq:bi-level}
\end{equation}
where $f_{\mathsf{up}}$ is the upper-level loss function, $f_{\mathsf{low}}$ is the lower-level objective that induces the UGNN, and $p_{pos}$ and $p_{pre}$ are the pre-processing and post-processing functions with parameters $\boldsymbol{\Theta}_{\mathsf{pre}}$ and $\boldsymbol{\Theta}_{\mathsf{pos}}$, respectively.
In the bi-level optimization problem \eqref{eq:bi-level}, the optimal solution of the lower-level problem $\bar{\mathbf{X}}\in\mathbb{R}^{N\times M^\prime}$ serves as input features to the upper-level objective with $M^\prime$ being the dimension of the node embedding.
These optimal features depend on learnable parameters of the lower-level problem in such a way that the entire bi-level pipeline can be trained end-to-end.

\begin{example}[Node classification with APPNP]
\label{example}
In an illustrative example, we use the APPNP model to solve a node classification problem.
The objective is to train an APPNP model on a graph $\mathcal{G}=\{\mathcal{V},\mathcal{E},\hat{\mathbf{X}},\mathbf{Y}\}$ to predict $\mathbf{Y}$ based on $\{\mathcal{V},\mathcal{E},\hat{\mathbf{X}}\}$.
In this scenario, we can choose $f_{up}$ as the cross-entropy loss function, $p_{\mathsf{pos}}$ as the $\mathrm{softmax}$ function, $f_\mathsf{low}$ as the GSD objective in Problem \eqref{eq:GSD}, and $p_{\mathsf{pre}}$ as a multi-layer perceptron.
\end{example}

Based on the optimization perspective of GNNs, a new design philosophy of GNNs has emerged, namely the UGNNs.
Unlike conventional GNNs that are often designed using heuristics, the forward layers of UGNNs are rigorously derived from the optimization steps for specific lower-level problems $f_low$.
The transparent nature of UGNNs allows us to develop GNNs in a more interpretable and controllable way.
The presence of an explicit optimization problem allows us to incorporate the desired inductive bias into UGNNs, for example, promoting the fairness of the model \cite{jiang2024chasing}, handling heterophily and heterogeneous graphs \citet{ahn2022descent,fu2022p}, and addressing potential attacks or noise in the node features or graph structure \cite{liu2021elastic,zhang2022asgnn,feng2023robust}, etc.
Furthermore, analyzing the obtained node embeddings based on their role in the $f_\mathsf{low}$ also gives us more explanatory details of the model \cite{zheng2024bloomgml}.

\section{Proposed Methodology}
\subsection{Design Motivation}\label{sec:motivation}
Through the lens of bi-level optimization, UGNNs inherently solve certain lower-level problems specified by $f_{\mathsf{low}}\left(\mathbf{H},\mathbf{L},p_{\mathsf{pre}}(\hat{\mathbf{X}},\boldsymbol{\Theta}_{\mathsf{pre}})\right)$.
Since $\mathbf{H}$ and $\boldsymbol{\boldsymbol{\Theta}_\mathsf{pre}}$ are parameters of the lower-level problem, the lower-level objective $f_{\mathsf{low}}$ is specified by $\mathbf{L}$ and $\mathbf{X}$.
Thus, if there is a distribution shift, i.e., transferring the model from the source domain to the target domain, the lower-level objective will change.

To investigate how $f_{low}$ changes during GDA, we design semi-supervised node classification experiments on two social networks, i.e., Twitch gamer networks collected in Germany (DE) and England (EN) \cite{rozemberczki2021multi}.
The experiments are performed with APPNP, whose corresponding $f_\mathsf{low}$ is defined as in Eq. \eqref{eq:GSD} and carries the meaning of GSD.
We examine how the value of $f_\mathsf{low}$ changes in different domains.
Specifically, we train an APPNP model on the source graph domain and then evaluate its corresponding $f_\mathsf{low}$ value on both the source domain and the target domains.
We report the average performance over 10 times in Table \ref{tab:GSD objective value}, where we use the min-max normalization to rescale the results to [0, 1].  

\begin{table}[ht]
\begin{centering}
\caption{GSD objective value corresponding to APPNP}
\label{tab:GSD objective value}
\par\end{centering}
\centering{}\resizebox{1.0\columnwidth}{!}{%
\begin{tabular}{ccccc}
\toprule 
\multirow{1}{*}{Source $\rightarrow$ Target} & DE $\rightarrow$ DE & EN $\rightarrow$ DE & EN $\rightarrow$ EN & DE $\rightarrow$ EN\tabularnewline
\midrule 
GSD objective & 0.1308 & 0.7687 & 0.3129 & 0.6812\tabularnewline
\bottomrule
\end{tabular}}
\end{table}
From the results, we observe that the GSD objective value $f_\mathsf{low}$ evaluated on the domain transfer setting, i.e., EN $\rightarrow$ DE and DE $\rightarrow$ EN are much larger than $f_\mathsf{low}$ evaluated on the in-domain setting, i.e, DE $\rightarrow$ DE and EN $\rightarrow$ EN.
Intuitively, as we transfer from the source to the target domain, the lower-level objective changes.
Thus, the model trained in the source domain cannot approximate the solution of the lower-level problem specified by the target domain $\mathcal{G}^t$ well, resulting in a large GSD objective value in the target domain.
We then naturally ask a question: \textit{Can we develop a strategy to maintain the level of noise of UGNNs measured by $f_\mathsf{low}$ while transferring across domains?}

\subsection{Domain Adaptive Unfolded Graph Neural Networks}
In this subsection, we analyze how $f_\mathsf{up}$ and $f_\mathsf{low}$ change during GDA, from the perspective of bi-level optimization. 
We focus on UGNNs, whose training process can be represented as in Eq. \ref{eq:bi-level}.
In the source domain, given $\mathbf{L}^s$, $\mathbf{X}^s$, and $\mathbf{Y}^s$, the UGNN model after training will have parameters
\begin{equation}
\begin{aligned}
\boldsymbol{\Theta}_{\mathsf{pre}}^{s},\boldsymbol{\Theta}_{\mathsf{pos}}^{s}\in&\mathrm{arg}\min_{\boldsymbol{\Theta}_{\mathsf{pre}},\boldsymbol{\Theta}_{\mathsf{pos}}}f_{\mathsf{up}}\Bigl(\mathbf{Y}^{s},p_{\mathsf{pos}}\Bigl(\mathrm{arg}\min_{\mathbf{H}}\\
&f_{\mathsf{low}}\left(\mathbf{H},\mathbf{L}^{s},p_{\mathsf{pre}}(\hat{\mathbf{X}}^{s},\boldsymbol{\Theta}_{\mathsf{pre}})\right),\boldsymbol{\Theta}_{\mathsf{pos}}\Bigr)\Bigr).
\end{aligned}
\label{eq:source model parameter}
\end{equation}
Similarly, we define an oracle model that is trained directly in the target domain $\mathcal{G}^{t}$ as
\begin{equation}
\label{eq:oracle}
\begin{aligned}
\boldsymbol{\Theta}_{\mathsf{pre}}^{\star},\boldsymbol{\Theta}_{\mathsf{pos}}^{\star}\in&\mathrm{arg}\min_{\boldsymbol{\Theta}_{\mathsf{pre}},\boldsymbol{\Theta}_{\mathsf{pos}}}f_{\mathsf{up}}\Bigl(\mathbf{Y}^{t},p_{\mathsf{pos}}\Bigl(\mathrm{arg}\min_{\mathbf{H}}\\
&f_{\mathsf{low}}\left(\mathbf{H},\mathbf{L}^{t},p_{\mathsf{pre}}(\hat{\mathbf{X}}^{t},\boldsymbol{\Theta}_{\mathsf{pre}})\right),\boldsymbol{\Theta}_{\mathsf{pos}}\Bigr)\Bigr).
\end{aligned}
\end{equation}
Then the goal of GDA becomes approximating $\boldsymbol{\Theta}_{\mathsf{pre}}^{\star},\boldsymbol{\Theta}_{\mathsf{pos}}^{\star}$ with $\boldsymbol{\Theta}_{\mathsf{pre}}^{s},\boldsymbol{\Theta}_{\mathsf{pos}}^{s}$.
The $p_{\mathsf{pos}}$ is normally chosen as parameter-free functions, e.g., the $\mathsf{softmax}$ function as in Example \ref{example}, or simple linear layers. 
To this account, we make the following assumption.
\begin{assumption}
\label{assu:pos_equality} The parameters in $p_\mathsf{pos}$ trained on the source graph domain and target graph domain are the same, i.e., $\boldsymbol{\Theta}_{\mathsf{pos}}^{s}=\boldsymbol{\Theta}_{\mathsf{pos}}^{\star}$.
\end{assumption}

Note that when $p_{\mathsf{pos}}$ is chosen as parameter-free functions, e.g., the $\mathsf{softmax}$ function, Assumption \ref{assu:pos_equality} always holds true.
When $p_{\mathsf{pos}}$ is chosen as linear layers, the reasoning ability still mainly lies in the lower-level problem part, indicating that Assumption \ref{assu:pos_equality} will not affect the expressivity of the model largely. 
Thus, a model under Assumption \ref{assu:pos_equality} should have similar performance compared with a model without Assumption  \ref{assu:pos_equality}. 
We provide an empirical validation of this point in Appendix.


Under Assumption \ref{assu:pos_equality}, when we apply the UGNN model trained in the source domain as defined in Eq. \eqref{eq:source model parameter} to the target domain, we will obtain the following loss value:
\[
f_{\mathsf{up}}\left(\mathbf{Y}^{t},p_{\mathsf{pos}}(\mathrm{arg}\min_{\mathbf{H}}f_{\mathsf{low}}\left(\mathbf{H},\mathbf{L}^{t},p_{\mathsf{pre}}(\hat{\mathbf{X}}^{t},\boldsymbol{\Theta}_{\mathsf{pre}}^{s})\right),\boldsymbol{\Theta}_{\mathsf{pos}}^{\star})\right).
\]
Compared to the loss function of the oracle model, the only difference is the embedding matrix $\bar{\mathbf{X}}$ obtained from the lower-level problem.
Thus, to make the loss value obtained with the UGNN model trained in the source domain close to the loss value obtained with the oracle model, we need the embedding matrix obtained by the UGNN model trained in source domain $\mathcal{G}^s$, i.e.,
\begin{equation}
    \bar{\mathbf{X}}^{s\rightarrow t}\in\mathrm{arg}\min_{\mathbf{H}}f_{\mathsf{low}}\left(\mathbf{H},\mathbf{L}^{t},p_{\mathsf{pre}}(\hat{\mathbf{X}}^{t},\boldsymbol{\Theta}_{\mathsf{pre}}^{s})\right)
\end{equation}
close to the embedding matrix obtained by the oracle model, i.e.,
\begin{equation}
\bar{\mathbf{X}}^\star\in\mathrm{arg}\min_{\mathbf{H}}f_{\mathsf{low}}\left(\mathbf{H},\mathbf{L}^{t},p_{\mathsf{pre}}(\hat{\mathbf{X}}^{t},\boldsymbol{\Theta}_{\mathsf{pre}}^{\star})\right).
\end{equation}
However, since we do not have access to $\mathbf{Y}^{t}$, we cannot obtain $\boldsymbol{\Theta}_{\mathsf{pre}}^{\star}$.
In other words, we do not have access to $\bar{\mathbf{X}}^\star$ during the training stage.
Thus, it is infeasible to directly make $\bar{\mathbf{X}}^{s\rightarrow t}$ close to $\bar{\mathbf{X}}^\star$.
To make the problem tractable, we instead try to make $f_\mathsf{low}^{s\rightarrow t}=f_{\mathsf{low}}\left(\bar{\mathbf{X}}^{s\rightarrow t},\mathbf{L}^{t},p_{\mathsf{pre}}(\hat{\mathbf{X}}^{t},\boldsymbol{\Theta}_{\mathsf{pre}}^{s})\right)$ close to $f_\mathsf{low}^\star=f_{\mathsf{low}}\left(\bar{\mathbf{X}}^\star,\mathbf{L}^{t},p_{\mathsf{pre}}(\hat{\mathbf{X}}^{t},\boldsymbol{\Theta}_{\mathsf{pre}}^{\star})\right)$.
However, the value of $f_\mathsf{low}^\star$ is still unavailable as it still relies on $\boldsymbol{\Theta}_{\mathsf{pre}}^{\star}$.
According to the empirical results in Section \ref{sec:motivation}, the value of $f_\mathsf{low}^\star$ is much smaller than the value of $f_\mathsf{low}^{s\rightarrow t}$ in practice.
Motivated by this observation, we suggest that simply reducing the value of $f_\mathsf{low}^{s\rightarrow t}$ can make it closer to $f_\mathsf{low}^\star$.

To achieve such a goal, we propose a simple yet effective strategy. Specifically, we propose to reset $p_{\mathsf{pre}}(\hat{\mathbf{X}}^{t},\boldsymbol{\Theta}_{\mathsf{pre}}^{s})$ as $\bar{\mathbf{X}}^{s\rightarrow t}$ and solve the lower-level problem again.
By doing so, we will obtain a new embedding matrix as follows:
\begin{equation}
\label{eq:reset}
\bar{\mathbf{X}}^{cp}\in\mathrm{arg}\min_{\mathbf{H}}f_{\mathsf{low}}\left(\mathbf{H},\mathbf{L}^{t},\bar{\mathbf{X}}^{s\rightarrow t}\right).
\end{equation}
In particular, a UGNN generates the solution to Eq. \eqref{eq:reset} by applying the message passing process with input node feature matrix $\bar{\mathbf{X}}^{s\rightarrow t}$ and graph Laplacian matrix $\mathbf{L}^t$.
In other words, this amounts to feeding the output of the message passing process in UGNN back to the model.
Thus, we name this strategy as cascaded propagation (CP).
Moreover, after CP, we will arrive at a new loss function value as
$f_\mathsf{low}^{cp}=f_{\mathsf{low}}\left(\bar{\mathbf{X}}^{cp},\mathbf{L}^{t},\bar{\mathbf{X}}^{s\rightarrow t}\right)$. 

\begin{assumption}
\label{assu:signal fidelity} The lower-level objective function $f_{\mathsf{low}}$ can be decomposed into three parts as
\[
\begin{aligned}
    f_{\mathsf{low}}&\left(\mathbf{H},\mathbf{L},p_{\mathsf{pre}}(\hat{\mathbf{X}},\boldsymbol{\Theta}_{\mathsf{pre}})\right)=\sum_{v\in\mathcal{V}}\kappa(\mathbf{h}_v,p_{\mathsf{pre}}(\hat{\mathbf{x}}_v,\boldsymbol{\Theta}_{\mathsf{pre}}))\\&\quad\quad\quad\quad\quad\quad\quad+\sum_{(u,v)\in\mathcal{E}}\xi\left(\mathbf{h}_u,\mathbf{h}_v \right)+\sum_{v\in\mathcal{V}}\eta(\mathbf{h}_v),
\end{aligned}
\]
where $\kappa$ represents the feature fidelity term, $\xi$ represents the feature smoothing term, and $\eta$ represents node-wise constraints.
Moreover, $p_\mathsf{pre}$ parameterized by $\boldsymbol{\Theta}$ only appears in function $\kappa$, which satisfies
\begin{equation}
    \kappa(\mathbf{h}_1,\mathbf{h}_2)\geq 0, \ \ \mathsf{and}\ \kappa(\mathbf{h}_1,\mathbf{h}_1)= 0, \ \ \forall \mathbf{h}_1, \mathbf{h}_2\in\mathbb{R}^{ M^\prime}.
\end{equation}
\end{assumption}

For UGNN models satisfying Assumption \ref{assu:signal fidelity}, after the CP process as defined in \eqref{eq:reset}, the value of the lower-level objective $f_\mathsf{low}$ is guaranteed to decrease, which is formally illustrated in the following theorem.
It is important to highlight that the bi-level optimization problems associated with most UGNNs satisfy Assumption \ref{assu:signal fidelity}, which also covers a wide spectrum of message-passing GNNs \cite{zheng2024bloomgml}.

\begin{theorem}\label{theorem}
Under Assumption \ref{assu:signal fidelity}, when transferring from the source domain to the target domain, the UGNNs with CP always have a smaller or equivalent lower-level objective value compared to their vanilla counterparts, i.e., 
\begin{equation}
\label{eq:Proof-goal}
\begin{aligned}
f_\mathsf{low}^{cp}&=f_{\mathsf{low}}\left(\bar{\mathbf{X}}^{cp},\mathbf{L}^{t},\bar{\mathbf{X}}^{s\rightarrow t}\right)\\ 
&\leq f_{\mathsf{low}}\left(\bar{\mathbf{X}}^{s\rightarrow t},\mathbf{L}^{t},p_{\mathsf{pre}}(\hat{\mathbf{X}}^{t},\boldsymbol{\Theta}_{\mathsf{pre}}^{s})\right). \\
&= f_{\mathsf{low}}^{s\rightarrow t}
\end{aligned}
\end{equation}
\end{theorem}

\begin{remark}
    According to Theorem \ref{theorem}, repeatedly applying CP can further reduce the lower objective. 
    However, UGNNs experience performance degradation when the number of layers is too high. 
    As a result, with additional CP, even though the gap in the performance after domain transfer narrows, the absolute performance does not necessarily improve.
\end{remark}

In the following, we elaborate in detail on how to integrate CP into UGNNs.
In particular, we employ CP with three representative UGNNs, namely APPNP \cite{gasteiger2019predict}, GPRGNN \cite{chien2021adaptive}, and ElasticGNN \cite{liu2021elastic}.
The GPRGNN model has the same lower-level objective as APPNP but with a learnable and layer-wise independent parameter $\alpha$.
The ElasticNet model is designed from the following lower-level problem:
\begin{equation}
\begin{aligned}
f_{\mathsf{low}}&\left(\mathbf{H},\mathbf{L},p_{\mathsf{pre}}(\hat{\mathbf{X}},\boldsymbol{\Theta}_{\mathsf{pre}})\right)=\frac{1}{2}\left\Vert \mathbf{H}-p_{\mathsf{pre}}(\hat{\mathbf{X}},\boldsymbol{\Theta}_{\mathsf{pre}})\right\Vert _{{\rm F}}^{2}\\&\quad\quad\quad\quad\quad\quad\quad+\frac{\lambda_1}{2}\mathrm{Tr}\bigl(\mathbf{H}^{\top}\mathbf{L}\mathbf{H}\bigr)+\lambda_2\left\Vert{\Delta}\mathbf{F}\right\Vert_1,
\end{aligned}
\end{equation}
where $\lambda_1$ and $\lambda_2$ are two weighting parameters and $\boldsymbol{\Delta}$ is the incident matrix satisfying $\boldsymbol{\Delta}^\mathsf{T}\boldsymbol{\Delta}=\mathbf{L}$.

\begin{example}[APPNP with CP]
The APPNP integrated with CP, named APPNP$_{CP}$, features the following message passing scheme:
\begin{equation}
\begin{aligned}
    \mathbf{H}^{(0)}=\mathbf{X},\ \ \ \ & \mathbf{H}^{(k+1)}=\left(1-\alpha\right)\mathbf{A}\mathbf{H}^{(k)}+\alpha\mathbf{X},\ \ \\& \text{for}\ k=0,\ldots,K-1, \\
&\mathbf{H}^{(k+1)}=\left(1-\alpha\right)\mathbf{A}\mathbf{H}^{(k)}+\alpha\mathbf{H}^{(K)},\ \ \\&\text{for}\ k=K,\ldots,2K,
\end{aligned}
\end{equation}
with $\mathbf{H}^{(2K)}$ being the output.
\end{example}

\begin{example}[GPRGNN with CP]
The GPRGNN integrated with CP, named GPRGNN$_{CP}$, has the following message passing scheme:
\begin{equation}
    \mathbf{H}^{(K)}=\sum_{k=0}^{K}\gamma^{(k)}\mathbf{A}^{k}\mathbf{X},\ \ \mathbf{H}^{(2K)}=\sum_{k=0}^{K}\gamma^{(k)}\mathbf{A}^{k}\mathbf{H}^{(K)},
\end{equation}
with $\mathbf{H}^{(2K)}$ being the output.
\end{example}

The message passing scheme of ElasticGNN integrated with CP, named ElasticGNN$_{CP}$, is provided in Appendix.
The CP strategy is an architectural enhancement for UGNNs, independent of information from the target domain, making it readily applicable to graph DG.
To further leverage the information from the target domain $\mathcal{G}^t=\{\mathcal{V}^t,\mathcal{E}^t,\mathbf{X}^t\}$ in GDA tasks, existing methods can be used to align the distribution of node representations between the source and target domains.
In our experiments, we incorporate a classical alignment loss, specifically MMD \cite{gretton2012kernel}, into the loss function with a trade-off parameter $\xi$.
The additional alignment loss helps to reduce domain discrepancies, facilitating the creation of domain-invariant representations.

The proposed CP strategy is compatible with UGNNs that can be represented in the form of Eq. \eqref{eq:bi-level}.
UGNNs can be flexibly designed to address node-level \cite{yang2021graph,han2023alternately,xue2023lazygnn}, edge-level \cite{wang2023efficient}, and graph-level  \cite{chen2022optimization} problems, allowing the CP strategy to improve the DG capabilities of UGNNs across various downstream tasks.
Furthermore, as an architectural enhancement, CP can complement existing UGDA methods that align node embedding distributions between the source and the target domains \cite{shi2024graph}.

\section{Experiments}
\label{sec:experiments}
In this section, we evaluate the effectiveness of the proposed CP strategy for UGNNs.
First, we introduce the experimental settings. 
Then, we assess the effectiveness of CP on node classification GDA tasks using three UGNN models: APPNP, GPRGNN, and ElasticGNN.
Finally, we conduct ablation studies to investigate the contributions of different components in UGNNs with CP.

\begin{table*}
\begin{centering}
\caption{Unsupervised GDA performance on citation networks.}
\label{tab:citation}
\par\end{centering}
\centering{}\resizebox{1.0 \textwidth}{!}{%
\begin{tabular}{ccccccccccccc}
\toprule 
 & \multicolumn{2}{c}{D $\rightarrow$ A}  & \multicolumn{2}{c}{C $\rightarrow$ A} & \multicolumn{2}{c}{A $\rightarrow$ D} & \multicolumn{2}{c}{C $\rightarrow$ D} & \multicolumn{2}{c}{A $\rightarrow$ C} & \multicolumn{2}{c}{D $\rightarrow$ C} \tabularnewline

 & Ma-F1 & Mi-F1 & Ma-F1 & Mi-F1 & Ma-F1 & Mi-F1 & Ma-F1 & Mi-F1 & Ma-F1 & Mi-F1 & Ma-F1 & Mi-F1\tabularnewline
 \midrule 
 DeepWalk & 19.83 & 26.23 & 19.33 & 21.94 & 19.87 & 25.94 & 17.51 & 22.57 & 17.72 & 21.05 & 22.76 & 29.46\tabularnewline
 Node2vec & 22.05 & 28.61 & 17.99& 21.76 & 19.50 & 24.54 & 24.98 & 28.95 & 25.84 & 29.89 & 16.22 & 21.16\tabularnewline

GCN & 59.42 & 63.35 & 70.39& 70.58 &  65.29 & 69.05&  71.37   &  74.53 &  74.78 & 77.38  & 69.79  & 74.17 \tabularnewline
GAT & 43.95 &  52.93 &  42.14 & 50.37 &  41.36  & 53.80  & 45.25  & 55.85&  43.64   & 57.13  & 50.04  & 55.52\tabularnewline
GIN & 56.50  &58.98& 59.48 & 60.46 & 50.49 &  59.10 & 63.48 & 66.27 & 62.49  & 68.61 &  63.21  & 69.25\tabularnewline

UDAGCN & 55.89 & 58.16 & 67.22 & 66.80 & 64.83 & 66.95 & 69.46 & 71.77 & 60.33 & 72.15 & 61.12 & 73.28  \tabularnewline
ACDNE & 72.64 & 71.29 & 74.79 & 73.59 & 73.59 & 76.24 & 75.74 & 77.21 & 80.09 & 81.75 & 78.83 & 80.14 \tabularnewline
ASN & 71.49 & 70.15 & 73.17 & 72.74 & 71.40 & 73.80 & 73.98 & 76.36 & 77.81 & 80.64 & 75.17 & 78.23 \tabularnewline
AdaGCN & 69.47 & 69.67 & 70.77 & 71.67 & 71.39 & 75.04 & 72.34 & 75.59 & 76.51 & 79.32 & 74.22 & 78.20  \tabularnewline
GRADE & 59.35 & 63.72 & 69.34 & 69.55 & 63.03 & 68.22 &  70.02 & 73.95 & 72.52 & 76.04 & 69.32 & 74.32  \tabularnewline
SpecReg & 72.34 & 71.01 & 73.15 & 72.04 & 73.98 & 75.93 & 73.64 & 75.74 &  78.83 & 80.55 & 77.78  & 79.04  \tabularnewline

A2GNN & 75.00 & 73.69 & 76.53 & 75.19 & 70.28 & 75.83 & 73.12 & 75.82 & 78.77 & 81.53 & 76.31 & 80.30\tabularnewline
 \midrule 
APPNP$_{CP}$ & 73.28 & 73.30 & 75.11 & 74.48 & \textcolor{brown}{\textbf{75.71}} & \textcolor{brown}{\textbf{77.80}} & \textcolor{brown}{\textbf{75.91}} & \textcolor{brown}{\textbf{77.77}} & 81.06 & 82.82 & 78.29 & 80.53\tabularnewline
GPRGNN$_{CP}$ & 74.78 & 74.50 & 75.52 & 74.71 & 75.15 & 77.42 & 75.04 & 77.55 & 80.95 & \textcolor{brown}{\textbf{82.83}} & \textcolor{brown}{\textbf{79.39}} & \textcolor{brown}{\textbf{81.02}}\tabularnewline
ElasticGNN$_{CP}$ & \textcolor{brown}{\textbf{75.74}} & \textcolor{brown}{\textbf{74.66}} & \textcolor{brown}{\textbf{76.72}} & \textcolor{brown}{\textbf{75.70}} & 75.49 & 77.38 & 75.78 & 77.64 & \textcolor{brown}{\textbf{81.16}} & 82.56 & 78.44 & 80.09\tabularnewline
\bottomrule
\end{tabular}}
\end{table*}

\subsection{Experiment Settings}
\textbf{Datasets.}
We conduct experiments on three citation networks, namely ACMv9 (A), Citationv1 (C), and DBLPv7 (D) \cite{zhang2021adversarial}, and two social networks, namely Germany (DE) and England (EN) \cite{rozemberczki2021twitch}.
Details on datasets are provided in Appendix.

\textbf{Baselines.}
We apply the proposed CP strategy to three UGNNs, named APPNP$_{CP}$, GPRGNN$_{CP}$, and ElasticGNN$_{CP}$. 
We use the MMD alignment loss to reduce domain discrepancy.
Note that the proposed CP method is compatible with other alignment losses as well. 
To evaluate the effectiveness of CP, we compared with several baselines, including node embedding methods: DeepWalk \cite{perozzi2014deepwalk} and Node2vec \cite{grover2016node2vec}; source only GNNs: GCN \cite{kipf2017semi}, graph attention network (GAT) \cite{velivckovic2018graph}, and graph isomorphism network (GIN) \cite{xu2019powerful}, which are only trained on the source graph without any DA techniques; and GDA methods: unsupervised domain adaptive GCNs (UDAGCN) \cite{wu2020comprehensive}, adversarial
cross-network deep network embedding (ACDNE) \cite{shen2020adversarial}, adversarial separation network (ASN) \cite{zhang2021adversarial}, adversarial domain
adaptation with GCN (AdaGCN) \cite{dai2022graph}, generic graph adaptive network (GRADE) \cite{wu2023non}, GNN with spectral regularization (SpecReg) \cite{you2023graph}, and asymmetric adaptive GNN (A2GNN) \cite{liu2024rethinking}. 
It is worth noting that our method enhances the models from an architectural perspective, which is perpendicular to most existing GDA methods.
Thus, they can be applied together to further enhance performance.

\textbf{Implementation Details.}
We follow the experiment setting introduced in \cite{wu2020unsupervised,shen2020adversarial} and applied as well in \cite{liu2024rethinking}.
We use 80\% of labeled nodes in the source domain for training, 20\% of the labeled nodes in the source domain for validation, and all the nodes in the unlabeled target domain for testing.
All the experiments are conducted on a Tesla V100 GPU.
The node representation dimension is set to 128, and the number of layers is set to 8.
We use the Adam optimizer \cite{KingBa15}.
We apply grid search for the learning rate and the weight decay parameter in the range of \{1e-4, 5e-4,1e-3, 5e-3\}.
The trade-off parameter $\xi$ for the MMD loss is searched in the range of \{1,2,3,4,5\}.
For APPNP and GPRGNN, there is an additional teleport parameter $\alpha$ which is searched in the range of \{0.1, 0.2, 0.5\}.
For ElasticGNN, there are two additional parameters $\lambda_1$ and $\lambda_2$ which are searched in the range of \{3,6,9\}.
The experiments are conducted five times using different seeds, and we report the average performance in terms of Macro-F1 (Ma-F1) and Micro-F1 (Mi-F1) scores. 
The Macro-F1 score represents the unweighted average of the  F1 scores for each class, treating all classes equally regardless of their sizes. 
In contrast,  the Micro-F1 score is the weighted average of the F1 scores for each class, where the weights correspond to the proportion of instances of each class in the dataset.

\subsection{Graph Domain Adaptation}

We evaluate the GDA performance of UGNNs with CP and compare them with other baselines.
We conduct six GDA tasks with citation networks, including A $\rightarrow$ C, A $\rightarrow$ D, C $\rightarrow$ A, C $\rightarrow$ D, D $\rightarrow$ A, D $\rightarrow$ C.
The results on these citation networks are presented in Table \ref{tab:citation}.
The best result is highlighted in bold and brown.
As shown in Table \ref{tab:citation}, UGNNs with CP achieve the best results on all GDA tasks.
On average, compared with the best result of the baselines on each GDA task, UGNNs with CP provide a 0.74\% improvement in the Macro-F1 score and a 0.92\% improvement in the Micro-F1 score.
Individually, each UGNN with CP also outperforms the best baseline on average.
Specifically, compared with the baseline that achieves the best average result, APPNP$_{CP}$ provides a 0.62\% improvement in the Macro-F1 score and a 0.72\% improvement in the Micro-F1 score; 
\begin{table}
\begin{centering}
\caption{Unsupervised GDA on social networks.}
\label{tab:social}
\end{centering}

\centering{}\resizebox{1.0 \columnwidth}{!}{%
\begin{tabular}{ccccc}
\toprule 
 & \multicolumn{2}{c}{DE $\rightarrow$ EN}  & \multicolumn{2}{c}{EN $\rightarrow$ DE}  \tabularnewline

 & Ma-F1 & Mi-F1 & Ma-F1 & Mi-F1 \tabularnewline
 \midrule 
 DeepWalk & 46.54 & 52.18 & 49.97 & 55.08\tabularnewline
 Node2vec & 46.96 & 52.64 & 50.10 & 54.61\tabularnewline

GCN & 54.55 & 54.77 & 51.04 & 62.03\tabularnewline
GAT & 49.50 & 54.84 & 40.08 & 43.65\tabularnewline
GIN & 49.91 & 52.39 & 44.26 & 55.26\tabularnewline

UDAGCN & \textcolor{brown}{\textbf{58.19}} & 59.74 & 56.35 & 58.69\tabularnewline
ACDNE & 56.31 & 58.08 & 57.92 & 58.79\tabularnewline
ASN & 51.21 & 55.45 & 45.90 & 60.45\tabularnewline
AdaGCN & 35.30 & 54.56 & 31.18& 40.22\tabularnewline
GRADE & 56.38 & 56.40 & 56.83 & 61.18\tabularnewline
SpecReg & 50.30 & 56.43 & 46.13 & 61.45\tabularnewline

A2GNN &  56.57 & 57.26 & 60.32 & 62.44\tabularnewline
 \midrule 
APPNP$_{CP}$ & 55.98 & 56.74 &  60.48 & 62.76 \tabularnewline
GPRGNN$_{CP}$ &  54.77 &  56.12 & 59.30  & 61.48\tabularnewline
ElasticNet$_{CP}$ & 57.69 & \textcolor{brown}{\textbf{59.87}} &  \textcolor{brown}{\textbf{61.39}} & \textcolor{brown}{\textbf{63.08}} \tabularnewline 
\bottomrule
\end{tabular}}

\end{table}
\begin{table*}
\begin{centering}
\caption{Performance contribution of MMD and CP for UGNNs.}
\label{tab:ablation}
\par\end{centering}
\centering{}\resizebox{1 \textwidth}{!}{%
\begin{tabular}{rcccccccccccc}
\toprule 
 \multirow{2}{*}{Method}& \multicolumn{2}{c}{D $\rightarrow$ A}  & \multicolumn{2}{c}{C $\rightarrow$ A} & \multicolumn{2}{c}{A $\rightarrow$ D} & \multicolumn{2}{c}{C $\rightarrow$ D} & \multicolumn{2}{c}{A $\rightarrow$ C} & \multicolumn{2}{c}{D $\rightarrow$ C} \tabularnewline

 & Ma-F1 & Mi-F1 & Ma-F1 & Mi-F1 & Ma-F1 & Mi-F1 & Ma-F1 & Mi-F1 & Ma-F1 & Mi-F1 & Ma-F1 & Mi-F1\tabularnewline
 \midrule 
APPNP & 63.46&  67.92& 72.15 & 72.54 & 63.91 & 68.47 & 74.20 & 76.77 & 76.30 & 79.05 & 71.87 & 75.38 \tabularnewline
+MMD & 72.20 &  72.07 & 73.96 & 73.92 & 73.92 & 76.90 & 75.41 & 77.53 & 80.36  & 82.19 &  76.61 & 78.68 \tabularnewline
+CP & 73.28 & 73.30 & 75.11 & 74.48 & 75.71 & 77.80 & 75.91 & 77.77 & 81.06 & 82.82 & 78.29 & 80.53\tabularnewline
\midrule 
GPRGNN & 61.44 & 68.58 & 73.52 & 73.47 & 63.08 & 67.43 & 73.63 & 76.93 & 77.71 & 80.57 & 73.10 & 77.21 \tabularnewline
+MMD & 70.92 & 72.27 & 74.89 & 74.14 & 74.30 & 76.64 & 74.18 & 77.26 & 80.14 & 82.69 & 77.80 & 79.95 \tabularnewline
+CP & 74.78 & 74.50 & 75.52 & 74.71 & 75.15 & 77.42 & 75.04 & 77.55 & 80.95 & 82.83 & 79.39 & 81.02\tabularnewline
\midrule
ElasticGNN & 65.51 & 68.82 & 71.83 & 72.05 & 62.94 & 66.22 & 72.10 & 74.31 & 75.33 & 77.75 & 74.22 & 76.97\tabularnewline
+MMD & 74.49 & 73.55 & 73.16 & 72.63 & 73.84 & 75.80 & 74.72 & 76.29 & 80.30 & 81.57 & 77.54 & 79.10\tabularnewline
+CP & 75.74 & 74.66 & 76.72 & 75.70 & 75.49 & 77.38 & 75.78 & 77.64 & 81.16 & 82.56 & 78.44 & 80.09\tabularnewline
\bottomrule
\end{tabular}}
\end{table*}
GPRGNN$_{CP}$ provides a 0.86\% improvement in the Macro-F1 score and a 0.94\% improvement in the Micro-F1 score; ElasticGNN$_{CP}$ provides a 1.34\% improvement in the Macro-F1 score and a 0.94\% improvement in the Micro-F1 score.

We also draw similar conclusions regarding social networks.
Specifically, we conduct two GDA tasks with social networks: DE $\rightarrow$ EN and EN $\rightarrow$ DE.
The node classification performance under GDA is presented in Table \ref{tab:social}.
On average, compared with the best result of the baselines on each GDA task, ElasticGNN$_{CP}$ provides a 0.28\% improvement in the Macro-F1 score and a 0.38\% improvement in the Micro-F1 score.
Notably, although UDAGCN performs well on the DE $\rightarrow$ EN task, ElasticGNN$_{CP}$ significantly outperforms it on the EN $\rightarrow$ DE task.
Specifically, ElasticGNN$_{CP}$  outperforms UDAGCN on the EN $\rightarrow$ DE task by 5.04\% in the Macro-F1 score and by 4.39\% in the Micro-F1 score.
In conclusion, UGNNs with CP can achieve better or comparable results compared with state-of-the-art GDA methods.

\subsection{Ablation Study}
In this section, we conduct ablation studies to investigate the role of different components in UGNNs with CP.
We provide results for APPNP, GPRGNN, and ElasticGNN in Table \ref{tab:ablation}.
We evaluate the performance of vanilla UGNNs, UGNNs with MMD (denoted as +MMD), and UGNNs with CP and MMD (denoted as +CP).
The results demonstrate that vanilla UGNNs do not perform well on the target domain.
By incorporating the MMD alignment loss, the performance of UGNNs significantly improves, indicating the importance of reducing domain discrepancy.
Finally, the results of UGNNs with CP demonstrate that the CP strategy can further improve model performance.

\begin{figure}
    \centering
     \includegraphics[width=1.0\columnwidth]{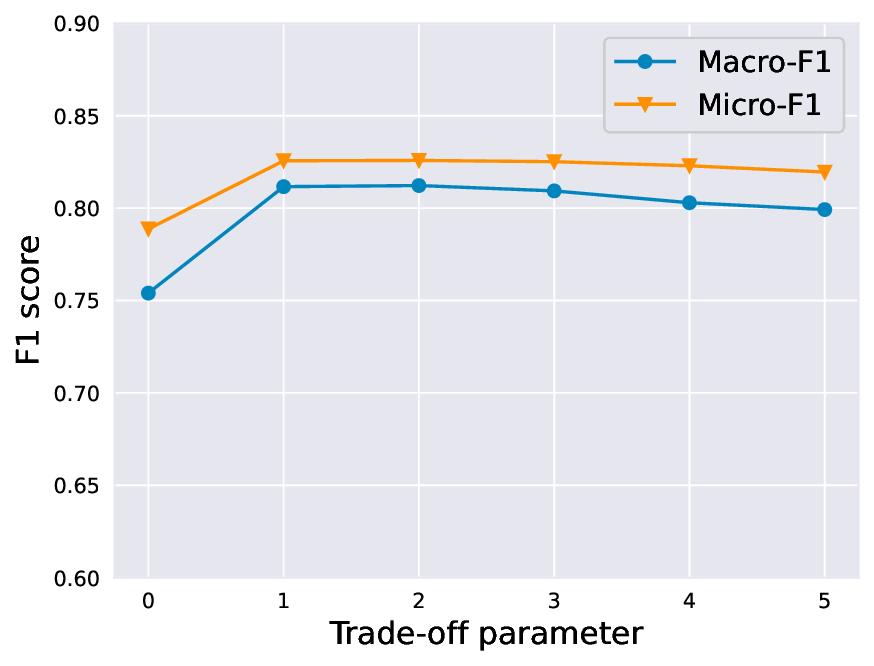}
    \caption{The influence of $\xi$ on A $\rightarrow$ C task.}
    \label{fig:trade-off-main}
\end{figure}
In UGNNs with CP, we apply the MMD as an additional loss, controlled by the trade-off parameter $\xi$.
In the following, we investigate the influence of the trade-off parameter $\xi$ on the performance of UGNNs with CP.
Specifically, we use the ElasticGNN$_{CP}$ as an example, as it performs the best in GDA according to the results in Table \ref{tab:citation} and Table \ref{tab:social}.
We conduct experiments on the GDA task A $\rightarrow$ C, varying the trade-off parameter $\xi$ from 0 to 5.
The result is visualized in Figure \ref{fig:trade-off-main}.
The results show that when $\xi=0$, the model does not perform well, highlighting the importance of minimizing domain discrepancy.
The model with MMD consistently outperforms the model without MMD for $\xi$ in the range of [1,5].
More ablation study results are provided in Appendix.

\section{Conclusion}
In this paper, we reveal through empirical and theoretical analyses that the lower-level objective function value associated with UGNNs significantly increases when transferring from the source domain to the target domain.
To address this, we propose a simple yet effective cascaded propagation strategy to enhance the domain generalization ability of UGNNs, which is provably effective in decreasing the lower-level objective function value in the target domain.
Experiments on five real-world datasets demonstrate the effectiveness of the cascaded propagation method.

The CP strategy is an architectural enhancement, and hence it can be used simultaneously with node representation distribution alignment methods.
The MMD alignment loss used in the experiments does not consider the specific characteristics of graph-structured data.
In future work,  alignment methods designed especially for graph-structure data can potentially improve performance.
Moreover, currently, the architectural enhancement (i.e., the proposed CP) and the alignment method (i.e., MMD) are developed independently; a more holistic design may lead to further performance improvement.
\newpage
\bibliography{ref}

\newpage
\appendix
\onecolumn
\section{Appendix}
\subsection{More Details on ElasticGNN with CP}\label{sec:elasticgnn_cp}
The ElasticNet model is designed based on the following lower-level objective:
\begin{equation}
f_{\mathsf{low}}\left(\mathbf{H},\mathbf{L},p_{\mathsf{pre}}(\mathbf{X},\boldsymbol{\Theta}_{\mathsf{pre}})\right)=\frac{1}{2}\left\Vert \mathbf{H}-p_{\mathsf{pre}}(\mathbf{X},\boldsymbol{\Theta}_{\mathsf{pre}})\right\Vert _{{\rm F}}^{2}+\frac{\lambda_1}{2}\mathrm{Tr}\bigl(\mathbf{H}^{\top}\mathbf{L}\mathbf{H}\bigr)+\lambda_2\left\Vert{\Delta}\mathbf{F}\right\Vert_1,
\end{equation}
where $\lambda_1$ and $\lambda_2$ are two weighting parameters and $\boldsymbol{\Delta}$ is the incident matrix.
The minimization of this lower-level objective can be achieved by solving an equivalent saddle point problem using the proximal alternating predictor-corrector algorithm \citep{loris2011generalization,chen2013primal}.
The optimization steps then induce a message passing scheme as follows:
\begin{equation}
\label{eq:emp}
\begin{cases}
\mathbf{Y}^{(k+1)} & =\gamma\mathbf{X}+(1-\gamma)\mathbf{A}\mathbf{H}^{(k)}\\
\mathbf{H}^{(k+1)} & =\mathbf{Y}^{(k)}-\gamma\boldsymbol{\Delta}^{\mathsf{T}}\mathbf{Z}^{(k)}\\
\mathbf{Z}^{(k+1)} & =\mathbf{Z}^{(k)}+\beta\boldsymbol{\Delta}\mathbf{H}^{(k+1)}\\
\mathbf{Z}_{i}^{(k+1)} & =\min\left(\left\Vert \mathbf{Z}_{i}^{(k+1)}\right\Vert _{2},\lambda_{1}\right)\frac{\mathbf{Z}_{i}^{(k+1)}}{\left\Vert \mathbf{Z}_{i}^{(k+1)}\right\Vert _{2}},\forall i\in[m]\\
\mathbf{H}^{(k+1)} & =\mathbf{Y}^{(k)}-\gamma\boldsymbol{\Delta}^{\mathsf{T}}\mathbf{Z}^{(k+1)}
\end{cases}
\mathsf{for} k=0,\ldots,K,
\end{equation}
where $\mathbf{H}^{(0)}=\mathbf{X}$ and $\mathbf{Z}^{(0)}=\mathbf{0}$.
With the CP, the ElasticGNN$_{CP}$ will further conduct the following computations subsequently to Eq. \eqref{eq:emp}:
\begin{equation}
\begin{cases}
\mathbf{Y}^{(k+1)} & =\gamma\mathbf{H}^{(K)}+(1-\gamma)\mathbf{A}\mathbf{H}^{(k)}\\
\mathbf{H}^{(k+1)} & =\mathbf{Y}^{(k)}-\gamma\boldsymbol{\Delta}^{\mathsf{T}}\mathbf{Z}^{(k)}\\
\mathbf{Z}^{(k+1)} & =\mathbf{Z}^{(k)}+\beta\boldsymbol{\Delta}\mathbf{H}^{(k+1)}\\
\mathbf{Z}_{i}^{(k+1)} & =\min\left(\left\Vert \mathbf{Z}_{i}^{(k+1)}\right\Vert _{2},\lambda_{1}\right)\frac{\mathbf{Z}_{i}^{(k+1)}}{\left\Vert \mathbf{Z}_{i}^{(k+1)}\right\Vert _{2}},\forall i\in[m]\\
\mathbf{H}^{(k+1)} & =\mathbf{Y}^{(k)}-\gamma\boldsymbol{\Delta}^{\mathsf{T}}\mathbf{Z}^{(k+1)}
\end{cases}
\mathsf{for} k=K,\ldots,2K,
\end{equation}
where $\mathbf{Z}^{(K)}=\mathbf{0}$.

\subsection{Additional Details on the Dataset} \label{sec:data_details}
We use three citation networks and two social networks in the experiments.
The three citation networks, namely ACMv9 (A), Citationv1 (C), and DBLPv7 (D) \citep{zhang2021adversarial}, are extracted from the original ArnetMiner dataset \citep{tang2008arnetminer}.
These networks are derived from different databases and over different time periods.
Specifically, ACMv9, Citationv1, and DBLPv7 are obtained from ACM (between 2000 and 2010), Microsoft Academic Graph (before 2008), and DBLP database (between 2004 and 2008), respectively.
Each network is represented as an undirected graph, with each node corresponding to a paper and each edge indicating a citation relation between two papers.
The task is to classify the paper according to their research topics.
The two social networks, namely Germany (DE) and England (EN), are the two largest networks in the Twitch gamer network dataset \citep{rozemberczki2021twitch}.
Each node represents a user, and each edge indicates a mutual follower relationship between the users.
The task is to classify the users based on whether they use explicit language.
The data statistics for these five datasets are summarized in Table \ref{tab:statistics}.
\begin{table}[h]
\begin{centering}
\caption{Dataset statistics.}
\label{tab:statistics}
\begin{tabular}{cccccc}
\hline 
Category & Datasets & \#Nodes & \#Edge & \#Attributes & \#Labels\tabularnewline
\hline 
\multirow{3}{*}{Citation} & DBLPv7 & 5484 & 8130 & 6775 & 5\tabularnewline
& ACMv9 & 9360 & 15602 & 6775 & 5\tabularnewline
& Citationv1 & 8935 & 15113 & 6775 & 5\tabularnewline
\hline 
\multirow{2}{*}{Social}& Garmany & 9498 & 153138 & 3170 & 2\tabularnewline
& England & 7126 & 35324 & 3170 & 2\tabularnewline
\hline 
\end{tabular}
\par\end{centering}
\end{table}

\subsection{Effect of the Weighting Parameter}\label{sec:more_ablation}
\begin{figure}[ht]

\centering
\begin{minipage}{0.4\textwidth}
\includegraphics[width=\textwidth]{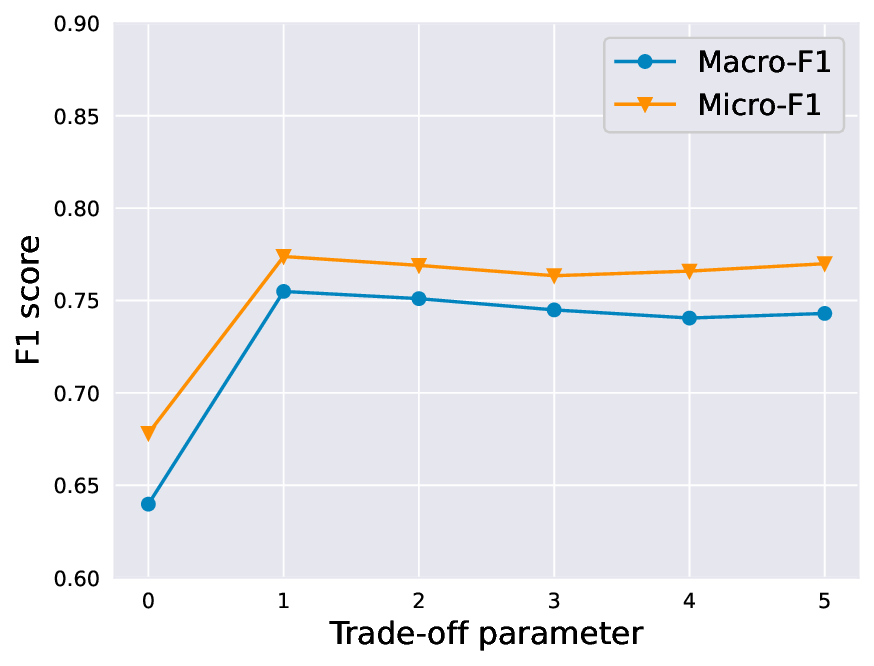}
\centerline{A$\rightarrow$D}
\end{minipage}
\begin{minipage}{0.4\textwidth}
\includegraphics[width=\textwidth]{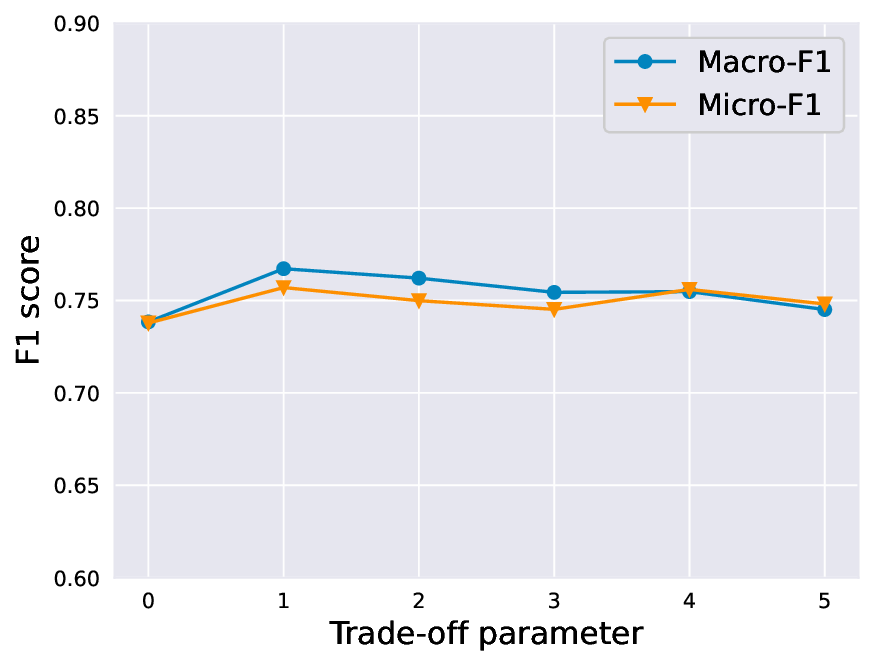}
\centerline{C$\rightarrow$A}
\end{minipage}

\begin{minipage}{0.4\textwidth}
\includegraphics[width=\textwidth]{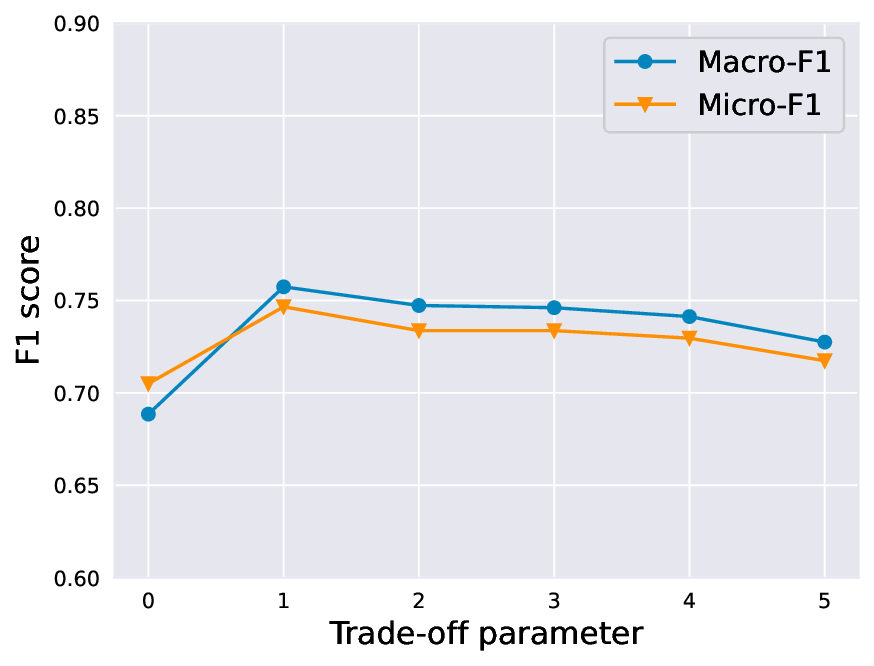}
\centerline{D$\rightarrow$A}
\end{minipage}
\begin{minipage}{0.4\textwidth}
\includegraphics[width=\textwidth]{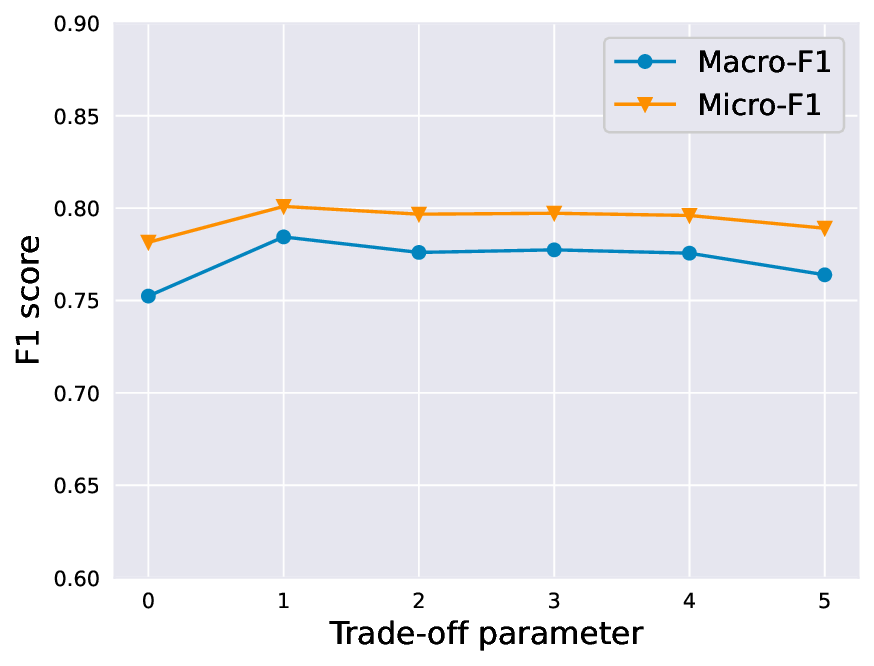}
\centerline{D$\rightarrow$C}
\end{minipage}
\caption{The influence of the trade-off parameter $\xi$ on ElasticGNN$_{CP}$.}\label{fig:weight parameter}
\end{figure}

In UGNNs with CP, we apply the MMD loss to minimize domain discrepancy, controlled by the trade-off parameter $\xi$.
In this section, we provide more results on evaluating the influence of the trade-off parameter $\xi$ on the performance of UGNNs with CP.
We conduct experiments on four additional GDA tasks in citation networks with ElasticGNN$_{CP}$, varying the trade-off parameter $\xi$ from 0 to 5.
The results are visualized in Figure \ref{fig:weight parameter}.
The results show that when $\xi=0$, the model does not perform well, highlighting the importance of minimizing domain discrepancy.
The model with MMD consistently outperforms the model without MMD for $\xi$ in the range of [1,5].
However, when $\xi$ becomes too large, the performance of ElsticGNN$_{CP}$ declines with increasing $\xi$. 
Therefore, an $\xi$ in the range of [1,2] is sufficient and yields the best results.

\subsection{Verification of Assumption 1}\label{sec:assump_verification}
\begin{figure}
\centering
\includegraphics[width=0.7\textwidth]{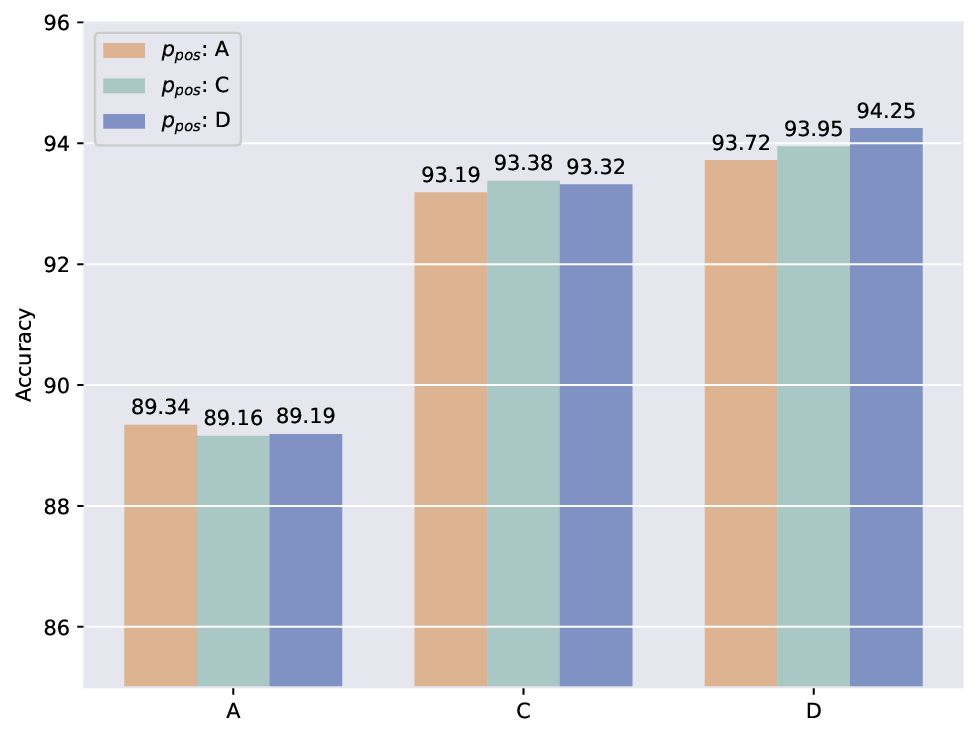}
\caption{Verification of Assumption \ref{assu:pos_equality}.}\label{fig:assumpr}
\end{figure}

In this section, we present empirical evidence using APPNP to demonstrate that Assumption \ref{assu:pos_equality} holds in practice.
Specifically, we begin by training the model on a source domain to obtain the parameters in the post-processing function $p_{pos}$.
We then train the model on a target domain with the parameters in $p_{pos}$ fixed as those obtained from the source domain.
We repeat this process for the six GDA tasks on citation networks, and the results in terms of node classification accuracy are reported in Figure \ref{fig:assumpr}, where A, C, and D represent dataset ACMv9, Citationv1, and DBLPv7, respectively. In the figure, the x-axis represents the target domain, and $p_{pos}:A$ indicates that the parameters in $p_{pos}$ are trained on dataset A.
For comparison, we also provide the results of the oracle model directly trained on the target domain data as detailed in Eq. \eqref{eq:oracle}, which indicates the upper bound of the performance.
From Figure \ref{fig:assumpr}, we conclude that the model trained on the target domain data with fixed $p_{pos}$ obtained from the source domain can achieve similar performance compared to the model trained on the target domain data without restrictions on $p_{pos}$.
Thus, the Assumption \ref{assu:pos_equality} is empirically validated.

\subsection{Proof of Theorem \ref{theorem}}
\begin{proof}
Given that $\bar{\mathbf{X}}^{cp}\in\mathrm{arg}\min_{\mathbf{H}}f_{\mathsf{low}}\left(\mathbf{H},\mathbf{L}^{t},\bar{\mathbf{X}}^{s\rightarrow t}\right)$, we have the following inequality:
\begin{equation}
f_\mathsf{low}^{cp}=
f_{\mathsf{low}}\left(\bar{\mathbf{X}}^{cp},\mathbf{L}^{t},\bar{\mathbf{X}}^{s\rightarrow t}\right)
\leq f_{\mathsf{low}}\left(\bar{\mathbf{X}}^{s\rightarrow t},\mathbf{L}^{t},\bar{\mathbf{X}}^{s\rightarrow t}\right).
\label{eq:proof-1}
\end{equation}
Under Assumption \ref{assu:signal fidelity}, which $f_\mathsf{low}$ satisfies, it can be expressed as
\begin{equation}
\begin{aligned}
    f_{\mathsf{low}}&\left(\bar{\mathbf{X}}^{s\rightarrow t},\mathbf{L}^{t},\bar{\mathbf{X}}^{s\rightarrow t}\right)
    =\sum_{v\in\mathcal{V}}\kappa(\bar{\mathbf{x}}^{s\rightarrow t}_v,\bar{\mathbf{x}}^{s\rightarrow t}_v)\\&+\sum_{(u,v)\in\mathcal{E}}\xi\left(\bar{\mathbf{x}}^{s\rightarrow t}_u,\bar{\mathbf{x}}^{s\rightarrow t}_v \right)+\sum_{v\in\mathcal{V}}\eta(\bar{\mathbf{x}}^{s\rightarrow t}_v).
\end{aligned}
\end{equation}
Since $\kappa(\mathbf{h}_1,\mathbf{h}_2)\geq 0$ and $\kappa(\mathbf{h}_1,\mathbf{h}_1)= 0$ for $\forall \mathbf{h}_1, \mathbf{h}_2\in\mathbb{R}^{ M^\prime}$, the following detailed calculation holds:
\begin{equation}
\begin{aligned}
\label{eq:Proof-2}
    &f_{\mathsf{low}}\left(\bar{\mathbf{X}}^{s\rightarrow t},\mathbf{L}^{t},\bar{\mathbf{X}}^{s\rightarrow t}\right)\\
    =&\sum_{(u,v)\in\mathcal{E}}\xi\left(\bar{\mathbf{x}}^{s\rightarrow t}_u,\bar{\mathbf{x}}^{s\rightarrow t}_v \right)+\sum_{v\in\mathcal{V}}\eta(\bar{\mathbf{x}}^{s\rightarrow t}_v),\\
    \leq &\sum_{v\in\mathcal{V}}\kappa(\bar{\mathbf{x}}^{s\rightarrow t}_v,p_{\mathsf{pre}}(\mathbf{x}^{t}_v,\boldsymbol{\Theta}_{\mathsf{pre}}^{s}))\\&+\sum_{(u,v)\in\mathcal{E}}\xi\left(\bar{\mathbf{x}}^{s\rightarrow t}_u,\bar{\mathbf{x}}^{s\rightarrow t}_v \right)+\sum_{v\in\mathcal{V}}\eta(\bar{\mathbf{x}}^{s\rightarrow t}_v)\\
    =&f_{\mathsf{low}}\left(\bar{\mathbf{X}}^{s\rightarrow t},\mathbf{L}^{t},p_{\mathsf{pre}}(\mathbf{X}^{t},\boldsymbol{\Theta}_{\mathsf{pre}}^{s})\right).
\end{aligned}
\end{equation}
Combining the result in Eq. \eqref{eq:proof-1} and Eq. \eqref{eq:Proof-2}, we arrive at the result presented in Eq. \eqref{eq:Proof-goal}.
\end{proof}
\end{document}